%% file: main.tex
\documentclass{article}
\usepackage{spconf,amsmath,graphicx,url}

\usepackage{graphicx,graphics}
\usepackage{epstopdf}
\DeclareGraphicsExtensions{.pdf,.png,.jpg,.eps}
\usepackage{xspace,xcolor}

\usepackage{amsmath}
\usepackage{amsthm}
\usepackage{amsfonts}

\usepackage{algorithm}
\usepackage{algorithmicx}
\usepackage[noend]{algpseudocode}

\input{math_definition}

\newcommand{\ibm}{\textsf{ibm}\xspace}
\newcommand{\rbm}{\textsf{rbm}\xspace}
\newcommand{\kernel}{\textsf{kernel}\xspace}

\title{A COMPARISON BETWEEN DEEP NEURAL NETS AND KERNEL ACOUSTIC MODELS FOR SPEECH RECOGNITION}
\name{Zhiyun Lu$^{1\dagger}$ \qquad Dong Guo$^{2\dagger}$ \qquad Alireza Bagheri Garakani$^{2\dagger}$ \qquad  Kuan Liu$^{2\dagger}$}
\aname{Avner May$^{3\ddagger}$ \qquad Aur\'{e}lien Bellet$^{4\ddagger}$ \qquad
Linxi Fan$^2$}
\bname{Michael Collins$^3$$^*$\thanks{$^*$Currently on leave at Google Inc. New York.}\qquad Brian Kingsbury$^5$ \qquad Michael Picheny$^5$ \qquad Fei Sha$^{1}$}
\address{
$^1$U. of California (Los Angeles)\quad  $^2$ U. of Southern California \quad $^3$Columbia U.\\
$^4$Team Magnet, INRIA Lille - Nord Europe
 \quad $^5$ IBM T. J. Watson Research Center (USA)
\\
{\small \emph{$^\dagger$\ $^\ddagger$: contributed equally as the first and  second co-authors, respectively}}\\
}

\begin{document}
\ninept
\sloppy
\maketitle
\begin{abstract}
\input{abs}
\end{abstract}
%

\input{intro}

\input{related}
\input{rks}

\input{results}
\input{discuss}
\input{ack}

\bibliographystyle{IEEEbib}
\bibliography{global_strings_learning,main,crossref}

\end{document}

%% file: math_definition.tex
\usepackage{amsmath,amsfonts}
\usepackage{amsthm}
\usepackage{amssymb,amsopn}
\usepackage{bm,bbm} 

\usepackage{amssymb}
\usepackage{amsmath,amsfonts}
\usepackage{amsthm,amsopn}

\usepackage{bm} 

\newcommand{\vct}[1]{\boldsymbol{#1}} 
\newcommand{\mat}[1]{\boldsymbol{#1}} 

\newcommand{\field}[1]{\mathbb{#1}}
\newcommand{\R}{\field{R}} 
\newcommand{\T}{^{\textrm T}} 

\newcommand{\ProbOpr}[1]{\mathbb{#1}}

\newcommand{\expect}[2]{%
\ifthenelse{\equal{#2}{}}{\ProbOpr{E}_{#1}}
{\ifthenelse{\equal{#1}{}}{\ProbOpr{E}\left[#2\right]}{\ProbOpr{E}_{#1}\left[#2\right]}}} 
\newcommand{\var}[2]{%
\ifthenelse{\equal{#2}{}}{\ProbOpr{VAR}_{#1}}
{\ifthenelse{\equal{#1}{}}{\ProbOpr{VAR}\left[#2\right]}{\ProbOpr{VAR}_{#1}\left[#2\right]}}} 

\DeclareMathOperator{\argmax}{arg\,max}

\newtheorem{theorem}{Theorem}

\newcommand{\vtheta}{\vct{\theta}}

\newcommand{\vx}{{\vct{x}}}

\newcommand{\vz}{{\vct{z}}}

\newcommand{\vphi}{\vct{\phi}}

\newcommand{\vdelta}{\vct{\delta}}
\newcommand{\vomega}{\vct{\omega}}

\newcommand{\eat}[1]{}

%% file: abs.tex
We study large-scale kernel methods for acoustic modeling and compare to DNNs on performance metrics related to both acoustic modeling and recognition. Measuring perplexity and frame-level classification accuracy, kernel-based acoustic models are as effective as their DNN counterparts. However, on token-error-rates DNN models can be significantly better. We have discovered that this might be attributed to DNN's unique strength in reducing both the perplexity and the entropy of the predicted posterior probabilities.  Motivated by our findings, we propose a new technique, entropy regularized perplexity, for model selection. This technique can noticeably improve the recognition performance of both types of models, and reduces the gap between them. While effective on Broadcast News, this technique could be  also applicable to other tasks. 

%% file: intro.tex
\section{Introduction}
\label{sIntro}

Deep neural networks (DNNs) have significantly advanced the state-of-the-art in automatic speech recognition (ASR)~\cite{bengio09deep,hinton12deep,mohamed12dbn,seide11dnn}.  In stark contrast, kernel methods, which had once been extensively studied due to their powerful modeling of highly nonlinear data~\cite{scholkopf02}, have not been competitive on large-scale ASR tasks.  There have been very few successful applications of kernel methods to ASR, let alone any ``head-on'' comparison to DNNs, except for a few efforts which were limited in scope~\cite{deng12convex,cheng11arcos,huang14kernel}. The most crucial challenge is that kernel methods scale poorly with the size of the training dataset, and thus are perceived as being impractical for ASR.


In this paper, we investigate empirically how kernel methods can be scaled up to tackle typical ASR tasks. We also study how they are similar to and different from DNNs. We focus on using kernel methods for frame-level acoustic modeling, but also evaluate them and contrast to DNNs on recognition performance. 

We have studied datasets for 3 languages and have made several interesting discoveries.  First, we show that kernel methods can tackle large-scale ASR tasks equally efficiently. To this end, we build on the random feature approximation technique, well-known in the machine learning community~\cite{rahimi07random}. Our contribution is to demonstrate its practical utility in constructing large-scale classifiers for acoustic modeling.  Second, we have found that kernel-based acoustic models are as good as DNN-based ones, {\em if their performance is measured in terms of perplexity or frame-level classification accuracy}.  However, when measuring word error rate (WER) performance, we have found that kernel-based acoustic models can lag significantly behind their DNN counterparts. For instance, on Broadcast News, IBM's DNN attains a WER of $16.7\%$ while the kernel-based model has $18.6\%$. There is a sharp difference despite the two having nearly identical frame-level perplexities.  Third, in the process of unraveling this mystery, we have discovered a new technique for selecting the best DNN acoustic model for decoding. Specifically, the new technique does \emph{not} stop training when the perplexity on the heldout data starts to worsen. Instead, it  looks at the tradeoff between the perplexity and the entropy of the \emph{predicted posterior probabilities} and favors models of lower entropy in exchange for a small sacrifice in perplexity.

Balancing these two factors leads to a new model selection criterion which we call \emph{entropy-regularized perplexity}. Acoustic models selected with it have better decoding results:  on the Broadcast News dataset the DNN WER improved to $16.1\%$ and the kernel model to $17.5\%$. We believe this criterion (and possible other variants) could be widely applicable for training DNNs for other ASR tasks.

The rest of the paper is organized as follows. We   review related work in \S\ref{sRelated}. Our empirical work focuses on scaling kernel methods up to large-scale problems -- we describe how in \S\ref{sRKS}.  In \S\ref{sResults}, we report extensive experiments comparing DNNs and kernel methods, followed by conclusions and discussion in \S\ref{sDiscuss}.

%% file: related.tex
\section{Related work}
\label{sRelated}

The computational complexity of exact kernel methods depends quadratically on the number of training examples at training time and linearly  at testing time. Hence, scaling up kernel methods has been a long-standing and actively studied problem~\cite{largescalekernelmachines07,smolaXX,decoste02invariant,platt98smo,tsang05cvm,clarkson10coreset}. Exploiting structures of the kernel matrix can scale kernel methods to 2 million to 50 million training samples~\cite{sonnenburg10coffin}.  

In theory, kernel methods provide a feature mapping to an infinite dimensional space. But, for any practical problem the dimensionality is bounded above by the number of training samples. Approximating kernels with finite-dimensional features has been recognized as a promising way of scaling up kernel methods. The most relevant approach for our paper is the observation~\cite{rahimi07random} that inner products between features derived from random projections can be used to approximate translation-invariant kernels~\cite{berg84,scholkopf02,rahimi07random}. Follow-up work on using those random features (``weighted random kitchen sinks''~\cite{rahimi08kitchen}) is a major inspiration for our work.  There has  been a growing interest in using random  projections to approximate different kernels~\cite{kar12random,hamid14compact,le13fastfood,vedaldi12additive}. 

Despite this progress,  there have been only a few reported large-scale empirical studies of those techniques on challenging tasks from speech recognition~\cite{deng12convex,cheng11arcos,huang14kernel}. However, the tasks were fairly small-scale (for instance, on the TIMIT dataset). By large, a thorough comparison to DNNs on ASR tasks  is lacking. Our work not only fills this gap, but also reveals details of the similarities and differences between those two popular learning paradigms.


%% file: rks.tex
\section{Kernel-based Acoustic Modeling}
\label{sRKS}
\subsection{Kernels and random features approximation}

Given a pair of data points $\vx$ and $\vz$, a positive definite kernel function $k(\cdot, \cdot): \R^d \times \R^d \rightarrow \R$ defines an inner product between the images of the two data points under a (nonlinear) mapping $\vphi(\cdot): \R^d \rightarrow \R^M$,
\begin{equation}
k(\vx, \vz) = \vphi(\vx)\T \vphi(\vz)
\label{eKernelDef}
\end{equation}
where the dimensionality $M$ of the resulting mapping $\vphi(\vx)$ can be infinite.  Kernel methods avoid inference in $\R^M$. Instead, they rely on the kernel matrix over the training samples. When $M$ is far greater than $N$, the number of training samples, this trick provides a nice computational advantage. However, when $N$ is exceedingly large, this quadratic complexity in $N$  becomes impractical.

\cite{rahimi07random} leverage a classical result in harmonic analysis and provide a fast way to approximate $k(\cdot, \cdot)$ with \emph{finite}-dimensional features:
\begin{theorem}{(Bochner's theorem, adapted from \cite{rahimi07random})}
A continuous kernel $k(\vx, \vz) = k (\vx - \vz)$ is positive definite if and only if $k(\vdelta)$ is the Fourier transform of a non-negative measure.
\end{theorem}
More specifically, for shift-invariant kernels such as Gaussian RBF and Laplacian kernels,
\begin{equation}
k^{\textsf{rbf}} = e^{-\|\vx-\vz\|_2^2/2\sigma^2},\quad k^{\textsf{lap}} = e^{-\|\vx-\vz\|_1/\sigma}
\label{eGauLap}
\end{equation}
 the theorem implies that the kernel function can be expanded with harmonic basis, namely
\begin{equation}
k(\vx - \vz)  = \int_{R^d} p(\vomega) e^{j\vomega\T(\vx-\vz)}\, d\vomega = \expect{\vomega}{e^{j\vomega\T\vx}e^{-j\vomega\T\vz}}
\label{eFourier}
\end{equation}
where $p(\vomega)$ is the density of a $d$-dimensional probability distribution. The expectation is computed on complex-valued functions of $\vx$ and $\vz$. For real-valued kernel functions, however, they can be simplified to the cosine and sine functions, see below. 

For Gaussian RBF and Laplacian kernels, the corresponding densities are Gaussian and Cauchy distributions:
\begin{equation}
\label{RBF}
p^{\textsf{rbf}}(\vomega) = N\left(0, \frac{1}{\sigma}\mat{I}\right),\quad p^{\textsf{lap}}(\vomega) = \prod_d \frac{1}{\pi(1+\sigma^2\omega_d^2)}
\end{equation}

This motivates a sampling-based approach of  approximating the kernel function. Concretely, we draw $\{\vomega_1, \vomega_2, \ldots, \vomega_D\}$ from the distribution $p(\vomega)$ and use the sample mean to approximate
\begin{equation}
k(\vx, \vz) \approx 1/D \sum_{i=1}^{D} \phi_{i}(\vx) \phi_{i}(\vz) =   \hat{\vphi}(\vx)\T\hat{\vphi}(\vz)
\label{eApprox}
\end{equation}
The \emph{\textbf{random feature vector}} $\hat{\vphi}$ is thus composed of scaled cosines of random projections
\begin{equation}
\hat{\phi}_{i}(\vx) =  \sqrt{2/D}\, \cos (\vomega_i\T\vx + b_i)
\label{eRandomFeature}
\end{equation}
where $b_i$ is a random variable, uniformly sampled from $[0,\ 2\pi]$. 

A key advantage of using approximate features  over standard kernel methods is its
scalability to large datasets. Learning with
a representation $\hat{\vphi}(\cdot) \in \R^D$ is relatively efficient
provided that $D$ is far less than the number of training samples.  For example, in our experiments (cf. section~\ref{sResults}), we have $7$ million to $16$ million  training samples, while $D \approx 25,000$ often leads to good performance.

\subsection{Use random features for acoustic modeling}

For acoustic modeling, we can plug the random feature vector $\hat{\vphi}(\vx)$ (converted from frame-level acoustic features) into a multinomial logistic regression model. Specifically, our model is a special instance of the \emph{weighted sum of random kitchen sinks}~\cite{rahimi08kitchen}
\begin{equation}
p( y = c |\vx) = \frac{e^{\vtheta_c\T \hat{\vphi}(\vx)}}{\sum_{c}e^{\vtheta_c\T \hat{\vphi}(\vx)}}
\label{eMLR}
\end{equation}
where the label $y$ can take any value from $\{1, 2, \ldots, C\}$, each corresponding to a phonetic state label. $\vtheta_c$ are learnable parameters.

\subsection{View kernel-acoustic model as a shallow neural network}

\begin{figure}
\centering
\includegraphics[width=0.5\columnwidth]{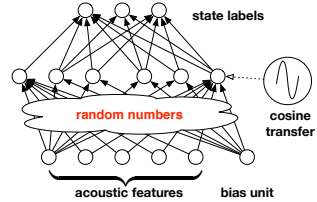}
\vspace{-1em}
\caption{Kernel-acoustic model seen as a shallow neural network}
\label{fRKS}
\vspace{-1em}
\end{figure}

The model eq.~(\ref{eMLR}) can be seen as a shallow neural network, shown in Fig.~\ref{fRKS}, with the following properties: (1) the parameters from the inputs (ie, acoustic feature vectors) to the hidden units are randomly chosen and not adapted; (2) the hidden units have $\cos(\cdot)$ as transfer functions; (3) the parameters from the hidden units to the output units are adapted (and can be optimized with convex optimization).

\subsection{Extensions}
The kernel acoustic model can also be extended to use the combination of multiple kernels -- graphically, they correspond to juxtaposing several shallow neural networks together~\cite{lu14arxiv}. 

The number of phonetic state labels can be very large. This will significantly increase the number of parameters in $\{\vtheta_c\}$. We can reduce it with a bottleneck layer (of 250 or 500 units) between the hidden units and the output layer. We experimented with two settings: a sigmoid bottleneck layer which corresponds to learning error-output-correct-code (ECOC) and a linear bottleneck layer which corresponds to low-rank factorization of the $\{\vtheta_c\}$~\cite{sainath2013low}.

%% file: results.tex
\section{Experimental Results}
\label{sResults}

We conduct extensive empirical studies comparing kernel methods to deep neural networks (DNNs) on typical ASR tasks.

\input{setup}

\subsection{Main Results}

\begin{table}
\centering
\caption{Comparison in perplexity (ppx) and accuracy (acc: \%)}
\vskip 0.5em
\label{tPerpAcc}
\begin{tabular}{|c|c|c|c|c|c|c|}\hline
& \multicolumn{2}{|c|}{Bengali} & \multicolumn{2}{|c|}{Cantonese} & \multicolumn{2}{|c|}{BN-50}  \\ \hline
Model   & ppx  & acc  & ppx  & acc   & ppx  & acc  \\ \hline \hline
DNN-\ibm  &   3.4  	&	71.5   	 	& 6.8 	&	56.8  	&  7.4 & 50.8\\ \hline
DNN-\rbm &  3.3  	& 	 72.1  	 	&  6.2  	&	 58.3   &	6.7 & 52.7 \\ \hline
\kernel &  3.5  &  71.0  & 6.5   	& 57.3   & 7.3& 51.2\\ \hline
\end{tabular}
\vskip -0.5em 
\end{table}

Table~\ref{tPerpAcc} contrasts the best  perplexity and accuracy attained by various acoustic models on held-outs. Note that cross-entropy errors (ie, the logarithm of the perplexity) are the training criteria of those models. Thus,  \textsf{ppx} correlates with classification accuracies  well.  Moreover, the performances by those 3 models are close to each other. Kernel models  have somewhat better performance than IBM's DNN.

\begin{table}
\small
\centering
\caption{Best token error rates on test set ($\%$)}
\label{tTER}
\vspace{0.5em}
\begin{tabular}{cccc}\hline
Model  & Bengali & Cantonese & BN-50\\ \hline
DNN-\ibm & 70.4  	&  	67.3 & 16.7	 \\ \hline
DNN-\rbm &  69.5 	& 	66.3  & 16.6\\ \hline
\kernel &  70.0	& 	 65.7 	& 18.6\\ \hline
\end{tabular}
\vskip -0.5em 
\end{table}

Table~\ref{tTER} reports the best \textsf{TER}s. On Bengali, \rbm performs marginally better than the kernel model, while \kernel performs  noticeably better than the two DNN models on Cantonese. However, the most surprising result is that \kernel performs significantly worse on BN-50 than either \rbm or \ibm. Note that in Table~\ref{tPerpAcc}, the kernel model has similar perplexity and accuracy as \rbm and better ones than \ibm. In what follows, we analyze the cause for this mismatch.

\subsection{Tradeoff between perplexity and entropy}

\begin{figure}
\centering
\includegraphics[width=0.75\columnwidth]{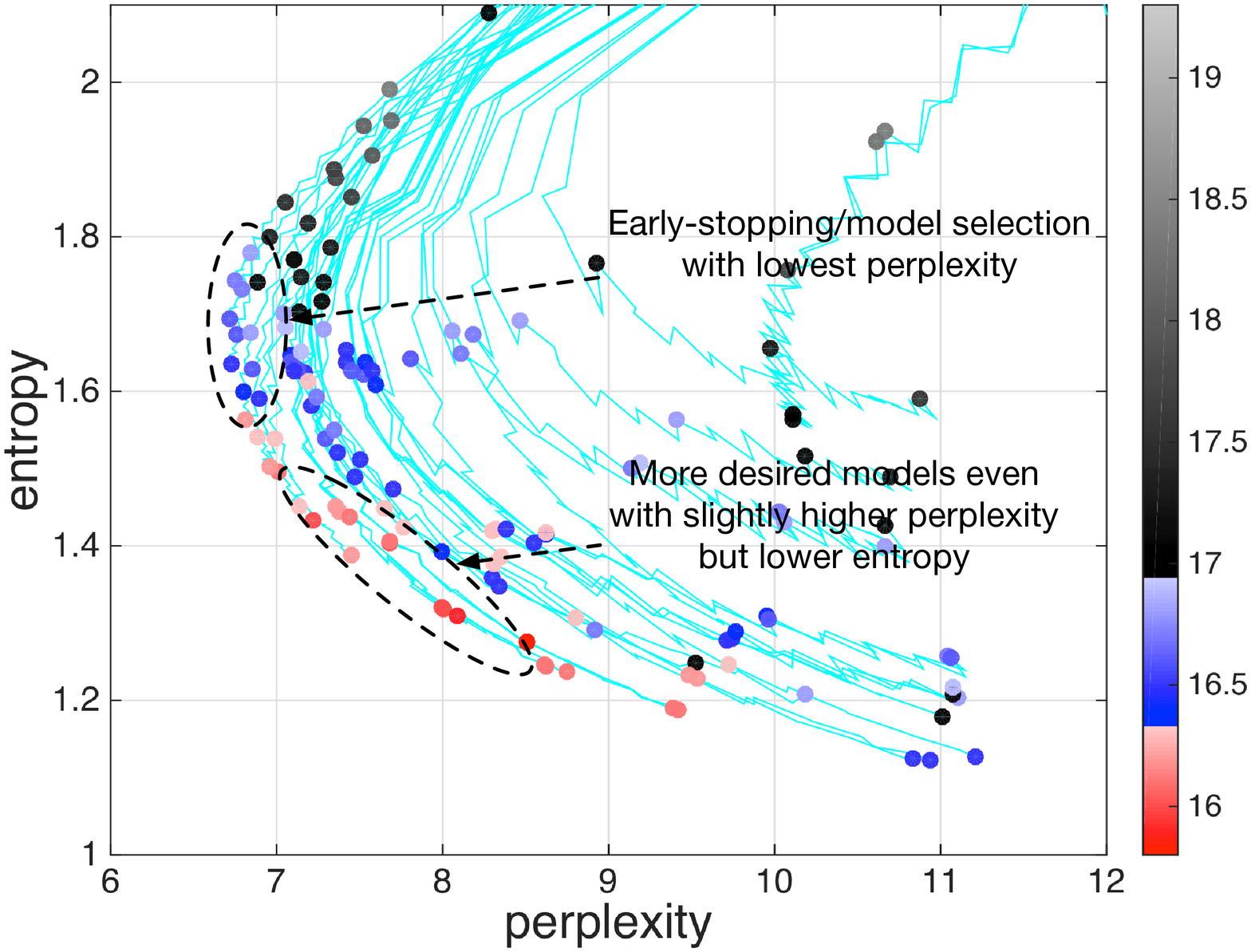}
\vskip -0.5em 
\caption{Training DNN acoustic models to minimize perplexity is not sufficient to arrive at the best WER -- after the typical early-stopping point where the perplexity is lowest, continuing to train to increase the perplexity but decrease the entropy leads to the best WER.}
\label{fDNNTrace}
\vskip -1em 
\end{figure}


One possible explanation is that the perplexity might  be an inadequate proxy for TER. As the predictions are probabilities to be combined with language models, we can capture the characteristics of the predictions using the entropy as it considers posterior probabilities assigned to all labels while the perplexity (as a training criteria) focuses only on the posterior probability assigned to the correct state label.  The entropy measures the degree of confusions in the predictions, which could have interplayed with the language models.

Fig.~\ref{fDNNTrace} plots the progression of several DNN models in perplexity and entropy (each cyan colored line corresponds to a model's training course). We also plot with colored markers the WERs evaluated at the end of every four epochs.  Clearly,  in the beginning of the training, both the entropy and the perplexity decrease, which also corresponds to an improving WER. Note that, using perplexity for early-stopping --- a common practice in training multinomial logistic regression model --- will result in models that are designated by the blue colored points on the leftmost of the plot. However, those models have sub-optimal WERs as continuing the training to have an increased perplexity but in exchange for a decreased entropy results in models with better WERs (the red colored points). 

\begin{figure}[t]
\centering
\includegraphics[width=0.75\columnwidth]{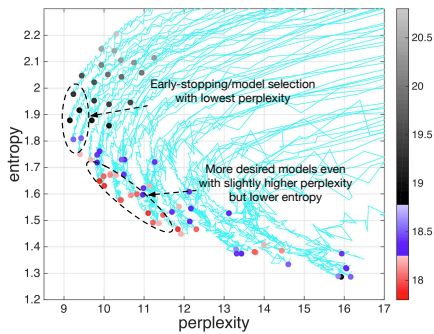}
\vskip -0.5em 
\caption{Similar to training DNN acoustic models, as  in Fig.~\ref{fDNNTrace}, training kernel models also has tradeoff between perplexity and entropy where the lowest perplexity does not correspond to the lowest WER. }
\label{fKernelSingleTrace}
\vskip -1em 
\end{figure}

We observe a similar tradeoff in training kernel-based acoustic models, as shown in Fig.~\ref{fKernelSingleTrace}. Similarly, WER depends jointly on the perplexity and the entropy and the best perplexity or entropy does not result in the best WER.  Note that when decoding, we tune the scaling of acoustic scores. Thus, balancing perplexity and entropy cannot be trivially achieved by scaling the inputs to the softmax.

\begin{figure}
\centering
\includegraphics[width=0.5\columnwidth]{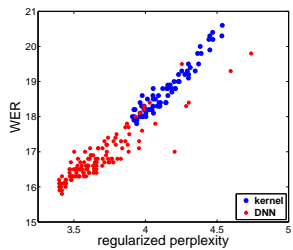}
\vskip -0.5em 
\caption{WER is almost linear in the regularized perplexity}
\label{fPredictTER}
\vskip -1em 
\end{figure}

\begin{table}[t]
\small
\centering
\caption{Regularized perplexity is a better model selection criteria}
\label{tRegularizedPPX}
\vspace{0.5em}
\begin{tabular}{cccc}\hline
Model  & perplexity & regularized perplexity & Oracle \\ \hline
\rbm &  16.6 	&  	16.1 & 15.8	  \\ \hline
\kernel & 18.6 & 17.5 & 17.5 \\ \hline 
\end{tabular}
\vskip -1em
\end{table}






\subsection{A new model selection criterion}

Fig.~\ref{fDNNTrace} and \ref{fKernelSingleTrace} suggest that we should select the best acoustic model using both perplexity and entropy.  Fig.~\ref{fPredictTER} shows that it is possible to predict WER from \emph{entropy-regularized perplexity}, defined as $\log(\text{perplexity})+\text{entropy}$, namely,
\begin{equation}
- \frac{1}{m}\sum_i \sum_{k=1}^K  [\mathbb{I}(k= y_i) +  P(y = k|x_i)]\log P(y=k|x_i)
\end{equation}
which has an almost linear relationship with the WER.

Table~\ref{tRegularizedPPX} illustrates the advantage of using this regularized perplexity on heldout to select models --  for both kernel and DNN acoustic models, their WERs are improved, and the improvement on kernel models is substantial ($1\%$ WER reduction in absolute).

While this new technique reduces the gap between kernel and DNN models, kernel method still lags behind. Continuing to pry is left to future work.

%% file: setup.tex
\subsection{Tasks, datasets and evaluation metrics}

We train both DNNs and kernel-based multinomial logistic regression models, as described in \S~\ref{sRKS}, to predict context-dependent HMM state labels from acoustic feature vectors. The acoustic features are 360-dimensional real-valued dense vectors, and are a standard speaker-adapted representation used by IBM~\cite{kingsbury13high}. The state labels are obtained via forced alignment using a GMM/HMM system.

We tested these models on three datasets.  The first two are the IARPA Babel Program Cantonese (IARPA-babel101-v0.4c) and Bengali (IARPA-babel103b-v0.4b) limited language packs.  Each pack contains a 20-hour training and a 20-hour test set.  We designate about 10\% of the training data as a held-out set to be used for model selection and tuning.  The training, held-out, and test sets contain different speakers.  Babel data is challenging because it is two-person conversations between people who know each other well (family and friends) recorded over telephone channels (in most cases with mobile telephones) from speakers in a wide variety of acoustic environments, including moving vehicles and public places.  As a result, it contains many natural phenomena such as mispronunciations, disfluencies, laughter, rapid speech, background noise, and channel variability.  An additional challenge in Babel is that the only data available for training language models is the acoustic transcripts, which are comparatively small.  The third dataset is a 50-hour subset of Broadcast News (BN-50)~\cite{kingsbury2009lattice,sainath2011making}.  45 hours of audio are used for training, 5 hours are a held-out set, and the test set is 2 hours.  This is well-studied benchmark task in the ASR community due to both its convenience and relevance to developing core ASR technology.


We use three metrics to evaluate the acoustic models:\\[1em]
\textsf{Perplexity} Given examples, \mbox{$\{ (\vx_i, y_i), i = 1 \ldots m\}$}, the perplexity is defined as $\textsf{ppx} = \exp\left\{-\frac{1}{m} \sum_{i=1}^m \log p(y_i | \vct{x}_i)\right\}$.  
Perplexity is usually correlated with the next two performance measures, so we use perplexity on the held-out for model selection and tuning.\\[0.5em]
\noindent
\textsf{Accuracy} The classification accuracy is defined as 
\[
 \textsf{acc} =  \frac{1}{m} \sum_{i=1}^m \mathbbm{1} \left[ y_i = \argmax_{y \in {1,2, \ldots, C}} p(y|\vct{x}_i)\right] 
\]
\textsf{Token Error Rate (TER)} We feed the predictions of acoustic models, which are real-valued probabilities, to the rest of the ASR pipeline and calculate the misalignment between the decoder's outputs and  the ground-truth transcriptions.  For Bengali and BN-50, the error is the word-error-rate (WER), while for Cantonese it is character error rate (CER). 

\subsection{Details of Acoustic Models}
For all kernel-based models, we use either Gaussian or Laplacian kernels and we also studied combinations of kernels.  For more details, please see~\cite{lu14arxiv}.  Kernel models have 3 hyperparameters: the bandwidths for Gaussian or Laplacian kernels, the number of random projections, and the step size of the (convex) optimization procedure (as adjusting it has a similar effect to early-stopping). As a rule of thumb, the kernel bandwidth ranges from 0.3--5 times the median of the pairwise distances in the data (with 1 times the median working well).  We typically use 2,000 to 400,000 random features, though stable performance is often observed at 25,000 or above.

For all DNNs, we tune hyperparameters related to both the architecture and the optimization. This includes the number of layers, the number of hidden units in each layer, the learning rate, the rate decay, the momentum, regularization, etc. We differentiate two types of DNNs: \ibm where the DNNs are first layer-wise discriminatively trained~\cite{seide11cddnn,kingsbury13high} and \rbm where the DNNs are first trained unsupervisedly and then discriminatively trained~\cite{hinton06dbn}.  \ibm is part of IBM's Attila package. For \rbm, we also tune hyperparameters for the unsupervised learning phase. 

\ibm acoustic model contains five hidden-layers, each of which contains 1024 units with logistic nonlinearities. The best \rbm has 4 hidden years, with 2000 hidden units per layer. The outputs of either types of models have either 1000 or 5000 softmax units, corresponding to the  quinphone context-dependent HMM states clustered using decision trees. All layers in the DNN are fully connected. For discriminative training, stochastic gradient descent with a mini-batch size of 250 samples, with tuned momentum, the learning rate annealing and early stopping on the held-outs.

%% file: discuss.tex
\section{Conclusion}
\label{sDiscuss}

As multiway classifiers, DNNs and kernel models do not seem to have significant differences when their performances are measured in terms of perplexity and accuracy. However, when integrated into the rest ASR pipeline, on Broadcast News (and possibly other) tasks, DNNs are able to attain much lower token error rates (TERs).  

Our analysis shows that when the perplexity and the entropy of the predicted posterior probabilities are balanced, models have  better TERs. Moreover, DNNs  can achieve lower entropy when they have similar perplexity as kernel models. Motivated by these findings, we design a ``regularized perplexity'' model selection/early-stopping criteria that select  better acoustic models which improve WERs over previous models that were selected  using un-regularized perplexity.

To the best of our knowledge, this paper is the first to pinpoint the unique niche  possessed by DNNs in better  integration with decoders. In future, we will try to understand why DNN has this appealing property despite being optimized with objectives  that do not take into consideration language models and structural loss~\cite{kingsbury2009lattice}.

%% file: ack.tex
\section{Acknowledgement}
This work is supported by the Intelligence Advanced Research Projects Activity (IARPA) via Department of Defense U.S. Army Research Laboratory (DoD/ARL) contract number W911NF-12-C-0012. The U.S. Government is authorized to reproduce and distribute reprints for Governmental purposes notwithstanding any copyright annotation thereon.  Disclaimer: The views and conclusions contained herein are those of the authors and should not be interpreted as necessarily representing the official policies or endorsements, either expressed or implied, of IARPA, DoD/ARL, or the U.S. Government.

Additional supports include USCÕs Center for High-Performance Computing (http://hpc.usc.edu), USC Provost Graduate Fellowship (ABG), NSF\#-1065243, 1451412, 1139148, a Google Research Award, an Alfred. P. Sloan Research Fellowship and ARO\# W911NF-12-1-0241 and W911NF-15-1-0484.